\documentclass[10pt, a4paper]{article}
\usepackage{lrec}
\usepackage{graphicx}
\usepackage{tabularx}
\usepackage{soul}

\usepackage{epstopdf}
\usepackage{inputenc}

\usepackage{hyperref}
\usepackage{xstring}

\title{Humor Detection in English-Hindi Code-Mixed Social Media Content :  Corpus and Baseline System}

\name{Ankush Khandelwal, Sahil Swami, Syed S. Akhtar,  Manish Shrivastava}

\address{\textit{Language Technology Research Centre, International Institute of Information Technology, Hyderabad} \\
         \{ankush.k, sahil.swami, syed.akhtar\}@research.iiit.ac.in\\
          m.shrivastava@iiit.ac.in}

\abstract{The tremendous amount of user generated data through social networking sites led to the gaining popularity of automatic text classification in the field of computational linguistics over the past  decade. Within this domain, one problem that has drawn the attention of many researchers is automatic humor detection in texts. In depth semantic understanding of the text is required to detect humor which makes the problem difficult to automate. With increase in the number of social media users, many multilingual speakers often interchange between languages while posting on social media which is called code-mixing. It introduces some challenges in the field of linguistic analysis of social media content \cite{barman2014code}, like spelling variations and non-grammatical structures in a sentence. Past researches include detecting puns in texts \cite{kao2016computational} and humor in one-lines \cite{mihalcea2010computational} in a single language, but with the tremendous amount of code-mixed data available online, there is a need to develop techniques which detects humor in code-mixed tweets. In this paper, we analyze the task of humor detection in texts and describe a freely available corpus containing English-Hindi code-mixed tweets annotated with humorous(\textbf{H}) or non-humorous(\textbf{N}) tags. We also tagged the words in the tweets with Language tags (\textbf{En}glish/\textbf{Hi}ndi/\textbf{Ot}hers). Moreover, we describe the experiments carried out on the corpus and provide a baseline classification system which distinguishes between humorous and non-humorous texts.
 \\ \newline \Keywords{humor detection, code-mixing, random forest classifier, SVM, extra tree classifier, naive bayes} }

\begin{document}

\maketitleabstract

\section{Introduction}

\begin{figure*}[h]
\begin{center}
\includegraphics[width=\textwidth]{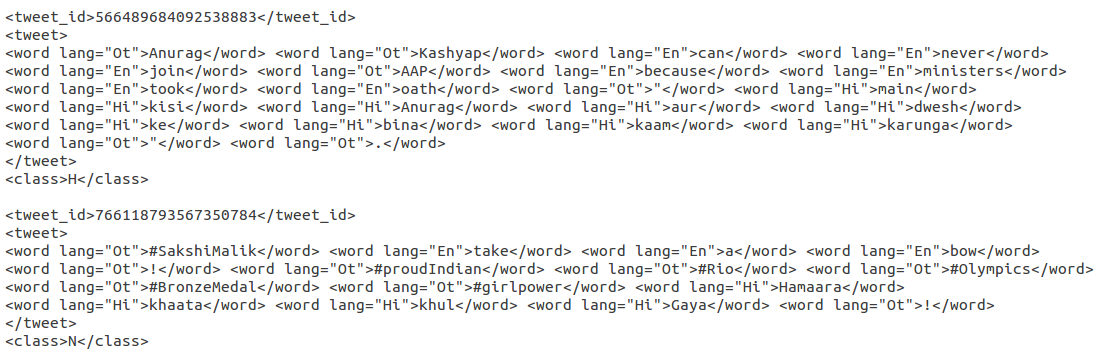} 
\caption{Example Annotations.}
\label{fig.1}
\end{center}
\end{figure*}

``Laughter is the best Medicine'' is a saying which is popular with most of the people. Humor is a form of communication that bridges the gap between various languages, cultures, ages and demographics. That's why humorous content with funny and witty hashtags are so much popular on social media. It is a very powerful tool to connect with the audience. 
Automatic Humor Recognition is the task of determining whether a text contains some level of humorous content or not. First conference on Computational humor was organized in 1996, since then many research have been done in this field. \newcite{kao2016computational} does pun detection in one-liners and \newcite{dehumor} detects humor in Yelp reviews. Because of the complex and interesting aspects involved in detecting humor in texts, it is one of the challenging research field in Natural Language Processing \cite{attardo1994linguistic}. Identifying humor in a sentence sometimes require a great amount of external knowledge to completely understand it. There are many types of humor, namely anecdotes, fantasy, insult, irony, jokes, quote, self deprecation etc \cite{hay1995gender,raz2012automatic}. Most of the times there are different meanings hidden inside a sentence which is grasped differently by individuals, making the task of humor identification difficult, which is why the development of a generalized algorithm to classify different type of humor is a challenging task.\\ 
Majority of the researches on social media texts is focused on English. A study by \newcite{schroeder2010half} shows that, a high percentage of these texts are in non-English languages. \newcite{fischer2011language} gives some interesting information about the languages used on Twitter based on the geographical locations. With a huge amount of such user generated data available on social media, there is a need to develop technologies for non-English languages. In multilingual regions like South Asia, majority of the social media users speak more than two languages. In India, Hindi is the most spoken language (spoken by 41\% of the population) and English is the official language of the country. Twitter has around 23.2 million monthly active users in India. Native speakers of Hindi often put English words in the sentences and transliterate the whole sentence to Latin script while posting on social media, thereby making the task of automatic text classification a very challenging problem. Linguists came up with a term for any type of language mixing, known as `code-mixing' or `code-switching'
 \cite{auer1999codeswitching,muysken2000bilingual,gafaranga2002interactional,bullock2011world}. Both the terms are used interchangeably, but there is a slight difference between the two terms. Code-mixing refers to the insertion of words, phrases, and morphemes of one language into a statement or an expression of another language, whereas transliteration of every word in a sentence to another script ( here Devanagari to Latin) is coined \textit{code-switching}\cite{alex2007automatic}. The first tweet in Figure 1 is an example of code-mixing and second is an example of code-switching. In this paper, we use code-mixing to denote both cases.\\
 In this paper, we present a freely available corpus containing code-mixed tweets in Hindi and English language with tweets written in Latin script. Tweets are manually classified into humorous and non-humorous classes. Moreover, each token in the tweets is also given a language tag which determines the source or origin language of the token (English or Hindi). 
The paper is divided in sections as follows, we start by describing the corpus and the annotation scheme in Section 2. Section 3 summarizes our supervised classification system which includes pre-processing of the tweets in the dataset and the feature extraction followed by the method used to identify humor in tweets. In the next subsection, we describe the classification model and the results of the experiments conducted using character and word level features. In the last section, we conclude the paper followed by future work and references.\\

\section{Corpus Creation and Annotation}

In this section, we explain the techniques used in the creation and annotation of the corpus. 

\subsection{Data Collection}

Python package twitterscraper\footnote{\url{https://github.com/taspinar/twitterscraper}}  is used to scrap tweets from twitter. 10,478 tweets from the past two years from domains like `sports', `politics', `entertainment' were extracted. Among those tweets, we manually removed the tweets which were written either in English or Hindi entirely. There were 4161 tweets written in English and 2774 written in Hindi. Finally, a total of 3543 English-Hindi code-mixed tweets were collected. Table 1 describes the number of tweets and words in each category.

\begin{table}[!h]
\begin{center}
\begin{tabularx}{\columnwidth}{|l|X|X|}

      \hline
      \textbf{Category} & \textbf{\#Tweets} & \textbf{\#Total Words}\\
      \hline
      Humorous (\textbf{H}) & 1755  & 26951 \\
      \hline
      Non-humorous (\textbf{N}) & 1698 & 21952\\
      \hline

\end{tabularx}
\caption{Description of the corpus.}
\end{center}
\end{table}

We tried to make the corpus balanced i.e. uniform distribution of tweets in each category to yield better supervised classification results as described by \cite{du2014supervised}. 

\subsection{Humor Annotation}
The final code-mixed tweets were forwarded to a group of three annotators who were university students and fluent in both English and Hindi. Approximately 60 hours were spent in tagging tweets for the presence of humor. Tweets which consisted of any anecdotes, fantasy, irony, jokes, insults were annotated as humorous whereas tweets stating any facts, dialogues or speech which did not contain amusement were put in non-humorous class. Following are some examples of code-mixed tweets in the corpus\footnote{Translation of Hindi words written in English are provided in the brackets. They are not a part of the tweet.}:
\begin{enumerate}
  \item ``For \#WontGiveItBack to work, Dhoni needs to say `Trophy toh ghar par hi bhul aaye' '' \\ \textit{(For \#WontGiveItBack to work, Dhoni needs to say ``We forgot the trophy at home'')} .
   \item Na sadak na naukri bas badh rahi gundagardi  \#FailedCMNitish
  \\ \textit{(No roads no naukri, only hooliganism increasing \#FailedCMNitish )}
  \item Subha ka bhula agar sham ko wapas ghar aa jaye then we must thank GPS technology.\\
  \textit{(If someone lost in the morning return home in the evening then we must thank GPS technology.)}
  \item She : Gori hai kalaiyan pehna de mujhe hari hari chudiya He *gets green bangles* She : Not this green ya, bottle green color.\\
  \textit{(She : (sings a song) My wrists are so beautiful, give me green green bangles He *gets green bangles* She : Not this green ya, bottle green color. )}
  \item Dard dilon ke kam ho jaate... twitter par agar poetries kam ho jaate \\ \textit{(pain in the heart will reduce...if number of poetries decreases on twitter)}.
\end{enumerate}
Annotators were given certain guidelines to decide whether a tweet was humorous or not. The context of the tweet could be found by searching about hashtag or keywords used in the tweet. Example (1) uses a hashtag `\#WontGiveItBack' which was trending during the ICC cricket world cup 2015. Searching it on Google gave 435k results and the time of the tweet was after the final match of the tournament. So there is an observational humor in (1) as India won the world cup in 2011 and lost in 2015 , hence the tweet was classified as humorous. Any tweets stating any facts, news or reality were classified as non-humorous. There were many tweets which did not contain any hashtags, to understand the context of such tweets annotators selected some keywords from the tweet and searched them online. Example (2) contains a comment towards a political leader towards development and was categorized as non-humorous. Tweets containing normal jokes and funny quotes like in (3) and (4) were put in humorous category. There were some tweets like (5) which consists of poem or lines of a song but modified. Annotators were guided that if such tweets contains satire or any humoristic features, then it could be categorized as humorous otherwise not. There were some tweets which were typical to categorize like (5), hence it was left to the annotators to the best of their understanding.
Based on the above guidelines annotators categorized the tweets. To measure inter annotator agreement we opted for Fleiss' Kappa \cite{fleiss1973equivalence} obtaining an agreement of 0.821 .   
Both humorous and non-humorous tweets in nearly balanced amount were selected to prepare the corpus. If we had included humorous tweets from one domain like sports and non humorous tweets from another domain like news then, it would have given high performance of classification \cite{dehumor}. To classify based on the semantics and not on the domain differences, we included both types of tweets from different domains. Many tweets contains a picture along with a caption. Sometimes a caption may not contain humor but combined with the picture, it can provide some degree of humor. Such tweets were removed from the corpus to make the corpus unimodal. In Figure 1, the first tweet, ``Anurag Kashyap can never join AAP because ministers took oath `main kisi Anurag aur dwesh ke bina kaam karunga' '' \textit{(Anurag Kashyap can never join AAP because ministers took oath `I will work without any affection (Anurag in Hindi) and without hesitation (dwesh in Hindi)')}, was classified as humorous. The second tweet, ``\#SakshiMalik take a bow! \#proudIndian \#Rio \#Olympics \#BronzeMedal \#girlpower Hamaara khaata khul Gaya!'' \textit{(\#SakshiMalik take a bow! \#proudIndian \#Rio \#Olympics \#BronzeMedal \#girlpower Our account opened!)} was classified as non-humorous as it contains a pride statement.   \\

\subsection{Language Annotation}

 Code-mixing provides some challenges for language identification like spelling variations and non-grammatical structures in a sentence  \cite{barman2014code}. Therefore, we annotated the tweets with the language at the word level. Native speakers of Hindi and proficient in English, labelled the language of the tokens in the tweets. Three types of tags were assigned to the tokens , \textbf{En} is assigned to the tokens present in English vocabulary like ``family'', ``Children'' etc. Similarly, \textbf{Hi} is assigned to the tokens present in Hindi vocabulary but transliterated to Latin script like ``samay'' (time), ``aamaadmi'' (common man). Rest of the tokens consists of proper nouns, numbers, dates, urls, hashtags, mentions, emojis and punctuations which are labelled as \textbf{Ot}(others). Major concern in language annotation was to annotate the words present in both languages (ambiguous words). For example, `to' (`but' in Hindi) and `is' (`this' in Hindi) , for this scenario annotators understood the context of the tweet and based on that the words were being annotated. For example, consider two sentences `This place is 500 years old' and `Is bar Modi Sarkar'\textit{(This time Modi Government)}. So `is' in first sentence should be tagged as English because it is used as a verb. It is transliterated to English in latter sentence and used as determiner (`this') so it was tagged as Hindi. So these kind of words were tagged based on their meanings and usage in both languages.
 Figure 1 describes the language annotation of tweets in the corpus.  
 
 \begin{table*}[ht]
\begin{center}

\begin{tabularx}{.8\textwidth}{|l|l|l|l|l|}

      \hline
      \textbf{Features (in \%)} & \textbf{Kernel SVM} & \textbf{Random Forest} & \textbf{Extra tree} & \textbf{Naive bayes}\\
      \hline
      N-grams & 68.5 & 63.7 & 65.4 & 68.2 \\
      \hline
      Bag-of-words & 60 & 61.6 & 61.6 & 67.3 \\
      \hline
       Common words and hashtags & 64.8 & 61.9 & 64.7 & 67.3\\
      \hline
       All features & \textbf{69.3}  & 65.2  & 67.8  & 67.2  \\
      \hline

\end{tabularx}
\caption{Accuracy of each feature using different classifiers}
 \end{center}
\end{table*}
 
\subsection{Annotation Scheme}

Two example annotations are illustrated in Figure 1. First line in every annotation consists of tweet id. Each tweet is enclosed within \textless tweet\textgreater\textless/tweet\textgreater~tags and each word is enclosed within \textless word lang=`` ''\textgreater \textless /word\textgreater~    containing the language annotation for each word. Last line determines the category in which the tweet belongs i.e. humorous or non-humorous, enclosed by \textless class\textgreater \textless /class\textgreater~tags. The annotated dataset with the classification system is made available online \footnote{ \url{https://github.com/Ankh2295/humor-detection-corpus}}.

\subsection{Error Analysis}
Social media users often make spelling mistakes or use multiple variations of a word while posting a text. We replaced all such errors with their correct version. For words in Hindi that were transliterated to Latin script, we adopted a common spelling for all those words across the corpus. For example, `dis' is often used as a short form for `this', so we replaced every occurrence of `dis' to `this' in the corpus. Some examples of spelling variations are, `pese' for `paisa' \textit{(unit of curreny)}, `h' for `hai' \textit{(is)}, `ad' for `add' \textit{(addition)}.

\section{System Architecture}

In this section, we describe our machine learning model which is trained and tested on the corpus described in the previous section. 

\subsection{Pre-processing of Corpus}

Preprocessing starts with tokenization, which involves separation of words using space as the delimiter and then converting the words to lower cases which is followed by the removal of punctuation marks from the tokenized tweet. All the hashtags, mentions and urls, are stored and converted to `hashtag', `mention' and `url' respectively. Sometimes hashtags provide some degree of humor in tweets, hence we segregated hashtags on the basis of camel cases and included the tokens in the tokenized tweets (hashtag decomposition) \cite{belainine2016named,ankushelection}. for example, \textit{\#AadabArzHai} can be decomposed into 3 words, `Aadab', `Arz' and `Hai'. Finally the tokenized tweets are stored along with the presence of humor as the target class.

\subsection{Classification Features}

 The features used to build attribute vectors for training our classification model are described below. We use character level and word level features for the classification \cite{ankushelection}. For all the features, we separated the words in the tweets based on the language annotation (Section 2.3) and prepared the feature vector for each tweet by combining the vectors for both the languages \footnote{Threshold values described are taken after empirical fine tuning}. 

\subsubsection{N-grams}

Previous researches shows that letter n-grams are very efficient for classifying text. They are language independent and does not require expensive text pre-processing techniques like tokenization, stemming and stop words removal, hence in the case of code-mix texts, this could yield good results \cite{miller2012gender,alowibdi2013language}. Since the number of n-grams can be very large we took trigrams which occur more than ten times in the corpus.

\subsubsection{Bag-of-words}

For classifying humor in texts, it is important to understand the semantics of the sentence. Thus, we took a three word window as a feature to train our classification models to incorporate the contextual information.

\subsubsection{Common Words and Hashtags}

 Many jokes and idioms sometimes have common words. We identified those words and took them as as a feature for classification. In the preprocessing step, we decomposed hashtags using camel cases and added them along with the words. Hence, common words in the hashtags were also included in the feature vector.

\subsection{Classification Approach and Results}

We experimented with four different classifiers, namely, support vector machine \cite{cristianini2000introduction}, random forest, extra tree and naive bayes classifier  \cite{pedregosa2011scikit}. Chi square feature selection algorithm is applied  to reduces the size of our feature vector. For training our system classifier, we used Scikit-learn \cite{pedregosa2011scikit}.\\
10-fold cross validation on 3543 code-mixed tweets was carried out by dividing the corpus into 10 equal parts with nine parts as training corpus and rest one for testing. Mean accuracy is calculated by taking the average of the accuracy obtained in each iteration of the testing process.
Table 2 shows the accuracy for each feature when trained using mentioned classifiers along with the accuracy when all the features are used along with the overall accuracy. Support vector machine with radial basis function kernel and extra tree classifier performs better than other classifiers and yields 69.3\% and 67.8\% accuracy respectively. The reason kernel SVM yields the best result is because the number of observations is greator than the number of features \cite{hsu2003practical}.   
N-grams proved to be the most efficient in all classification models followed by common words and hastags. Bag-of-words feature performed the worst in SVM, random forest and extra tree classifier but yielded better result in naive bayes classifiers. Accuracies mentioned in table 2 were calculated using fine tuning of model parameters using grid search.

%

\section{Conclusion and Future Work}
In this paper, we describe a freely available corpus of 3453 English-Hindi code-mixed tweets. The tweets are annotated with humorous(H) and non-humorous(N) tags along with the language tags at the word level. The task of humor identification in social media texts is analyzed as a classification problem and several machine learning classification models are used. The features used in our classification system are n-grams, bag-of-words, common words and hashtags. N-grams when trained with support vector machines with radial basis function kernel performed better than other features and yielded an accuracy of 68.5\%. The best accuracy (69.3\%) was given by support vector machines with radial basis function kernel.\\  
This paper describes the initial efforts in automatic humor detection in code-mixed social media texts. Corpus can be annotated with part-of-speech tags at the word level which may yield better results in language detection. Moreover, the dataset can be further extended to include tweets from other domains. Code-mixing is very common phenomenon on social media and it is prevalent mostly in multilingual regions. It would be interesting to experiment with code-mixed texts consisting of more than two languages in which the issue of transliteration exists like Arabic, Greek and South Asian languages. Comparing training with code-mixed tweets with training with a merged corpus of monolingual tweets in English and Hindi could be an interesting future work.

\section{Bibliographical References}
\label{main:ref}

\bibliography{xample} 
\bibliographystyle{lrec}

\label{lr:ref}

\end{document}